\documentclass[a4paper, 10pt, conference]{ieeeconf}      

\IEEEoverridecommandlockouts                              

\usepackage[caption=false]{subfig}
\usepackage{svg}
\usepackage{amsmath} 
\usepackage{amssymb}  
\usepackage{adjustbox}
\usepackage{multirow}

\title{\LARGE \bf
GAFAR: Graph-Attention Feature-Augmentation for Registration\\A Fast and Light-weight Point Set Registration Algorithm
}

\author{Ludwig Mohr$^{1}$, Ismail Geles${^1}$ and Friedrich Fraundorfer$^{1}$%
\thanks{$^{1}$Institute of Computer Graphics and Vision, Graz University of Technology, Inffeldgasse 16, 8010 Graz, Austria.
{\tt\footnotesize [ludwig.mohr, fraundorfer]@icg.tugraz.at}}%
}

\begin{document}

\maketitle
\thispagestyle{empty}
\pagestyle{empty}

\begin{abstract}
Rigid registration of point clouds is a fundamental problem in computer vision with many applications from 3D scene reconstruction to geometry capture and robotics.
If a suitable initial registration is available, conventional methods like ICP and its many variants can provide adequate solutions.
In absence of a suitable initialization and in the presence of a high outlier rate or in the case of small overlap though the task of rigid registration still presents great challenges.
The advent of deep learning in computer vision has brought new drive to research on this topic, since it provides the possibility to learn expressive feature-representations and provide one-shot estimates instead of depending on time-consuming iterations of conventional robust methods.
Yet, the rotation and permutation invariant nature of point clouds poses its own challenges to deep learning, resulting in loss of performance and low generalization capability due to sensitivity to outliers and characteristics of 3D scans not present during network training.

In this work, we present a novel fast and light-weight network architecture using the attention mechanism to augment point descriptors at inference time to optimally suit the registration task of the specific point clouds it is presented with.
Employing a fully-connected graph both within and between point clouds lets the network reason about the importance and reliability of points for registration, making our approach robust to outliers, low overlap and unseen data.
We test the performance of our registration algorithm on different registration and generalization tasks and provide information on runtime and resource consumption.
The code and trained weights are available at \verb|https://github.com/mordecaimalignatius/GAFAR/|.

\end{abstract}

\section{Introduction}

Rigid registration of point clouds is the task of simultaneously inferring both pose and correspondences between two sets of points~\cite{3d3d}.
As soon as either pose or correspondences are known, estimation of the respective other is straight forward, yet doing both simultaneously is posing challenges in computer vision and robotics.
Its importance in tasks such as pose estimation~\cite{pose-robotics,pose-pipe}, map-building and SLAM~\cite{RoboticsSLAM, Scan2Model} as well as localization tasks geared towards autonomous driving~\cite{stickyloc} fuel the research interest in registration algorithms.

ICP and its many variants~\cite{icp, icp-real}, while able to provide exceptional results for good initializations, tend to get stuck in local minima if the initialization is insufficient, in the presence of high outlier rates, or in cases with low overlap.
Attempts to resolve this range from methods using branch-and-bound to infer a globally optimal solution~\cite{GoICP}, methods based on feature matching between key-points followed by robust matching strategies~\cite{ransac,irls} and in recent years deep neural networks for learning feature descriptors and matching~\cite{FCGF,PPFNet,PointNetLK,DCP}.
Yet both, branch-and-bound as well as robust matching, suffer from speed and accuracy issues in real-life application due to the high number of iterations necessary in cases of high outlier ratios.
Deep-learning based methods usually fare better with regard to outliers, yet still struggle due to the contradiction between low distinctiveness of local point-features caused by topological similarities and low match recall of global features in low-overlap cases.
A further drawback of algorithms using deep neural networks often is their high requirements concerning compute resources, limiting their use in mobile applications.

\begin{figure}[t]
 \centering
 \vspace{15pt}
 \subfloat{\includegraphics[width=0.55\columnwidth]{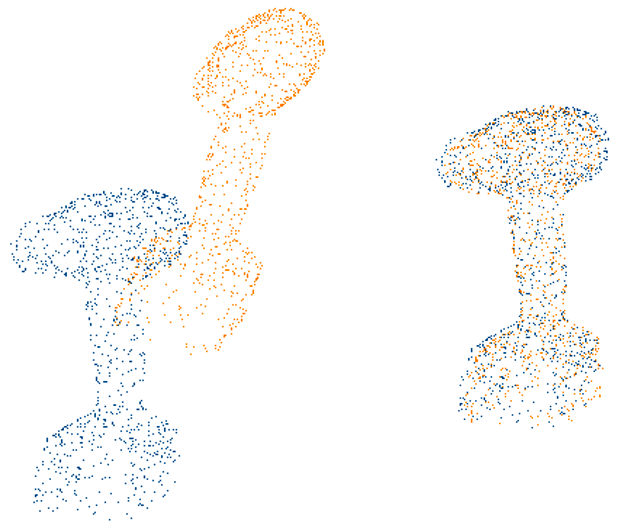}}{}
 \hfill
 ~
 \subfloat{\includegraphics[width=0.35\columnwidth]{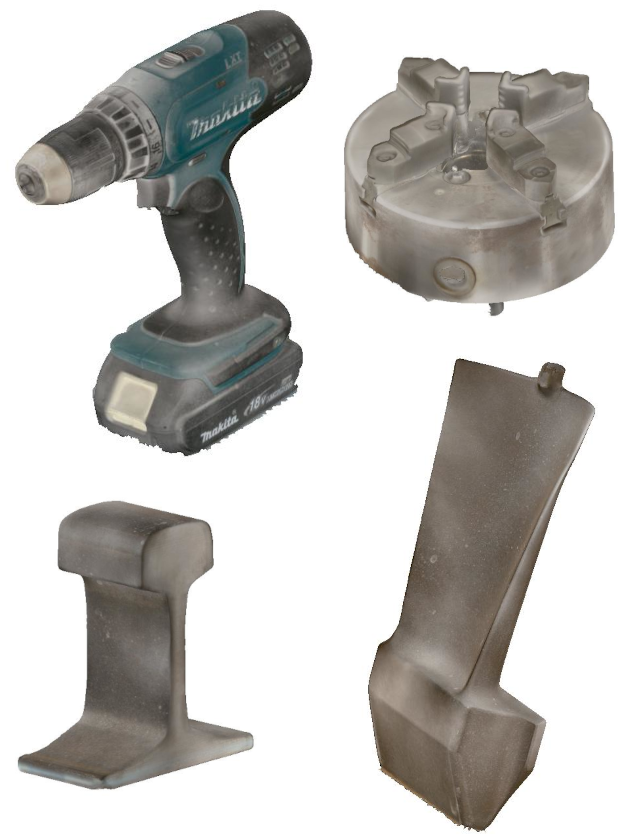}}{}
 \caption{Registration example on high quality real world scans captured with a handheld 3D scanner (left) and examples of the (meshed and textured) object scans available in the custom dataset used for testing generalization ability (right).}
 \label{fig:dmv}
\end{figure}

To tackle these challenges, we propose GAFAR: Graph-Attention Feature-Augmentation for Registration, which employs deep-learning techniques not only for extraction of meaningful local features from point sets, but also for learning an adaptive augmentation network for online transformation of local features for robust matching.
We achieve this by exploiting structural information from between point sets as well as from within a single one thorough an architecture of interleaved self- and cross-attention layers~\cite{attention, superglue}.
While achieving state-of-the-art registration performance, our method is light-weight and fast.

We demonstrate this in a series of experiments, testing not only registration performance on the dataset used for training, but also robustness and generalization ability in two further experiments on vastly different datasets of real world scans, one captured with a handheld 3D scanner producing precise scans, the other being the Kitti Odometry Dataset~\cite{kitti} showing street scenes captured by a LiDAR scanner.
Furthermore, we provide insight into runtime and resource needs.
The main contributions of our method are:
\begin{itemize}
 \item We demonstrate the use of transformer networks and the attention mechanism to build a fast and light-weight, yet accurate registration algorithm.
 \item We present an online feature augmentation strategy in registration which proves to be superior in terms of robustness to partial overlap and geometries not seen during training.
 \item We show how certain design-choices enable us to estimate the registration success without knowledge of the true transformation, enabling its use in applications that require fail-safes.
 \item We demonstrate state-of-the art performance and superior generalization capability in a light-weight package.
\end{itemize}

\section{Related Work}
One of the oldest, yet still relevant methods for registration of point clouds is ICP~\cite{icp-o3d}.
Starting from an initial alignment, ICP iteratively updates the registration parameters by establishing point correspondences using Euclidean distance, rejecting far away point pairs.
Due to this design it is prone to get stuck in local minima, the final registration accuracy heavily depends on the initialization.
Many variants have been proposed over the years~\cite{icp} to mitigate these issues, yet the dependence on the initialization has remained.

Several registration algorithms trying to solve the dependence on initialization have been proposed~\cite{GoICP, FGR}, alongside of handcrafted feature descriptors trying to capture local geometry of point clouds in a meaningful way, such as PPF~\cite{PPF} and FPFH~\cite{fpfh}, among others~\cite{3Dfeat}.
Yet, they never managed to reach the performance and robustness of their 2D counterparts.

Recent advances in deep-learning extend deep neural networks to 3D point clouds and have resulted in methods for learned local feature descriptors like PointNet~\cite{pnet,pnetpp}, FCGF~\cite{FCGF}, Graphite~\cite{graphite} and DGCNN~\cite{dgcnn}, learned filtering of putative point matches~\cite{DGR} and complete learned registration pipelines.
3DSmoothNet~\cite{3dsmooth} extracts a local reference frame and voxelizes the point cloud around key points, yet reference frame estimation is susceptible to outliers, voxelization tends to loose information due to spatial discretization.
PointNetLK~\cite{PointNetLK} estimates registration parameters to match the deep representations from PointNet of complete point clouds, DCP~\cite{DCP} uses DGCNN to extract point features and the attention mechanism~\cite{attention} to predict soft correspondences, restricting their application to registration of point clouds with high overlap.
Research into Pillar-Networks~\cite{stickyloc,stickypillars} is driven mainly by automotive applications for processing of LiDAR point clouds from mobile mapping systems, assuming the input point clouds to share a common z=up orientation.
They extract cylindrical point pillars along the z-axis around key point locations for further processing, and are therefor not applicable to general registration problems or when the assumption of z-axis alignment can not be guaranteed.
Keypoint based methods like~\cite{predator} aim at detecting repeatable keypoints across scans, and registering them using powerful descriptors.
In contrast~\cite{geotrans} uses a detection-free approach with a local-to-global detection strategy using superpoints.
IDAM~\cite{idam} tackles inaccuracies arising from inner product norms for feature matching with an iterative distance-aware similarity formulation.
DeepGMR~\cite{deepgmr} recovers registration parameters from Gaussian Mixture Models, parameterized using pose-invariant correspondences.
RPM-Net~\cite{RPMNet} predicts annealing parameters and predicts correspondences with annealing in feature matching and the Sinkhorn Algorithm~\cite{SinkhornOT} as solver for linear assignment, predicting soft correspondences.
RGM~\cite{RGM} explicitly builds and matches graphs within point clouds to resolve ambiguity issues between locally similar patches and predicts hard correspondences using the Hungarian Algorithm.

In contrast to ~\cite{RGM}, we use graph matching for feature augmentation before matching, but do not match graphs extracted from point clouds explicitly.
Similarity between the internal point cloud structures is handled by our method implicitly using cross-attention modules.
Our method predicts hard correspondences by thresholding of the assignment matrix after running sinkhorn iterations, interpreting the correspondence estimation as optimal transport problem of the feature correlation matrix.

\section{Problem Formulation}
Rigid registration of two 3D point sets is the task of finding a transformation consisting of a rotation matrix $\mathbf{R} \in \mathbf{SO}^3 $ and a translation vector $\mathbf{t} \in \mathbb{R}^3$ aligning input point set $\mathcal{P}_{S} = \{p_i \in \mathbb{R}^3 | i = 1, ..., M\}$ to the reference point set $\mathcal{P}_{R} = \{p_j \in \mathbb{R}^3 | j = 1, ..., N\}$.
Here $M$ and $N$ denote the respective sizes of the point sets.

The underlying assumption is, that both point sets are sampled on the same surface or the same object and share at least some common support (i.e., the physical location where the object has been sampled does actually overlap).
In the most general case, point sets $\mathcal{P}_S$ and $\mathcal{P}_R$ may not have any true correspondences between them, may suffer from outliers and additive noise and they may only share parts of their support, resulting in only partial overlap.

Given a set of corresponding points between two point sets, the rigid transformation aligning both sets can be recovered using SVD.
This approach relaxes the task of estimating a rigid transformation to that of finding pairs of corresponding points between both sets.
Since the transformation obtained using SVD aligns the point pairs in a least-squares sense, this formulation directly lends itself to the case where no exact matches exist.

Hence, the task of rigid point set registration can be formulated mathematically as:
\begin{equation}
 \mathbf{C}^* = \underset{\mathbf{C}}{\mathrm{arg min}} \biggl(\sum_j^N \sum_i^M c_{i,j} \Vert \mathbf{R}_Cp_i + \mathbf{t}_C - p_j \Vert^2 \biggr),
 \label{eq:perm}
\end{equation}
where $\mathbf{C} \in \{0, 1\}^{M,N}$ is a permutation matrix subject to row and column constraints $\sum_i^M C_i = \mathbf{1}^N$ and $\sum_j^N C_j = \mathbf{1}^M$, associating the points between both point sets.
The transformation parameters $\mathbf{R}_C$ and $\mathbf{t}_C$ refer to those recovered by SVD using the point pairs designated by permutation matrix $\mathbf{C}$.
To handle the case of partial overlap, the permutation matrix is augmented by a row and column to $C \in \{0, 1\}^{M+1,N+1}$, while relaxing the constraints on rows and columns of $\mathbf{C}$ to
\begin{equation}
 \sum_i^M C_i \leq \mathbf{1}^N,
 \hspace{15pt}
 \sum_j^N C_j \leq \mathbf{1}^M
 \label{eq:perm_relaxed}.
\end{equation}

In practice, this formulation can be solved by augmenting an initial full point feature correlation matrix with an additional row and column and solving the relaxed optimization problem as the optimal transport problem~\cite{hungarian, OT}, using the Sinkhorn Algorithm as differentiable implementation of the linear assignment problem~\cite{SinkhornOT, superglue}.

\section{The Making of GAFAR}
\begin{figure*}[thpb]
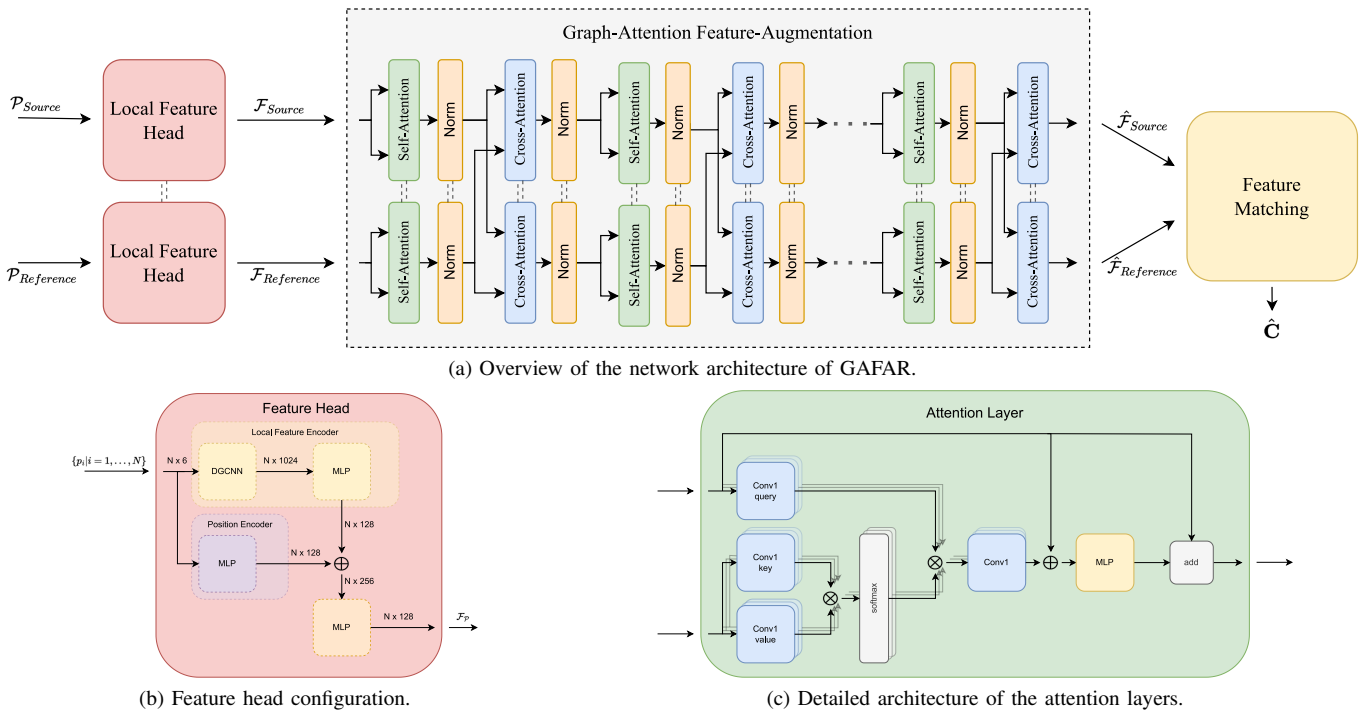

  \centering
  \subfloat[Overview of the network architecture of GAFAR.\label{fig:gafar}]{\includesvg[width=\textwidth]{./img/gafar_architecture.svg}}%
  \hfill
  \vspace{3pt}
  \subfloat[Feature head configuration.\label{fig:head}]{\includesvg[width=0.3\textwidth]{./img/feature_head.svg}}%
  \hspace{2cm}
  ~
  \subfloat[Detailed architecture of the attention layers.\label{fig:attention}]{\includesvg[width=0.47\textwidth]{./img/AttentionLayer.svg}}%
  \caption{
  Architectural details of GAFAR.
  Figure~(\ref{fig:gafar}) shows an overview of the network architecture, sub-figures~(\ref{fig:head}) and (\ref{fig:attention}) the structure of the feature head and the attention layers, respectively.
  $\bigotimes$ denotes matrix multiplication, $\bigoplus$ concatenation.}
  \label{fig:arch_all}
\end{figure*}

The key idea behind our network architecture is to adapt initial local per-point feature descriptors $\mathcal{F}_S$ of a source point set $\mathcal{P}_S$ for correspondence matching in an online fashion by injecting information of the reference point set $\mathcal{P}_R$.
The reasoning behind this is, that for successful point matching neither only local geometric structure (which may be repetitive or non-distinctive) nor fully-global information (which in case of partial overlap may encode information of areas which are not shared) is sufficient.
The relevant information for successful point matching lies solely within the topology of the overlapping area as well as the relative position of points within this area.
Our architecture takes two point sets $\mathcal{P}_S$ and $\mathcal{P}_R$, represented as point locations in Euclidean coordinates together with their respective point normals, as input.
Internally, the network architecture consists of a feature head generating per-point features for both point sets independently, as well as an augmentation stage inspired by~\cite{superglue}, consisting of interleaved self- and cross-attention layers.
This allows the network to reason jointly over both sets of feature descriptors, adapting them iteratively into representations optimally suited for finding high-quality correspondences between those two specific point sets.
Matching is done by calculating the dot-product similarity between all possible pairings of the resulting feature descriptors $\hat{\mathcal{F}}_S$ and $\hat{\mathcal{F}}_R$, relaxing the match matrix by adding a slack row and slack column and running the Sinkhorn Algorithm a predefined number of iterations, as in~\cite{superglue,RPMNet}.
The network weights are shared between the two branches processing $\mathcal{P}_S$ and $\mathcal{P}_R$, turning the architecture into a fully-siamese network~\cite{siamese}.
Figure~\ref{fig:gafar} depicts an overview of the architecture, the different building blocks are explained in greater detail in the following subsections.

\subsection{Local Feature Descriptor Head}
Our feature head, depicted in Figure~\ref{fig:head}, consists of two main building blocks, a local feature encoder with a neighbourhood size $\mathcal{N}$ and a point-wise location encoder Multi-Layer Perceptron (MLP).
Both take point locations within the unit-circle and their respective normal vectors as input.
As point-feature network we employ an architecture derived from DGCNN~\cite{dgcnn}, extended by an MLP functioning as a bottleneck to reduce the feature dimensionality to a more suitable size.
A basic layer of this architecture embeds the lower-dimensional representation into a higher dimensional local representation with a nonlinear transformation by applying a MLP on point patches consisting of the $\mathcal{N}$ nearest neighbours of each point $p_i$, followed by max-pooling over the patch and normalization.
Information is aggregated via multiple layers and concatenation until a high-dimensional internal representation $\mathcal{F}_I \in \mathbb{R}^d$ of the local point neighbourhood is reached.
Our point-wise location encoder is implemented as a pure point-wise MLP, for each point $p$ in point set $\mathcal{P}$ embedding its position in Euclidean space into a high-dimensional feature space, again of size $\mathbb{R}^d$.
The output of both, the feature encoder and the position encoder, are then concatenated and projected back to $\mathbb{R}^d$ by a small point-wise MLP.

\subsection{Graph-Attention Feature-Augmentation Network}
The purpose of the graph attention network for feature augmentation is to optimize the feature representations $\mathcal{F}_S$ of the input point set $\mathcal{P}_S$ at inference time for correspondence search by infusing knowledge of the reference point set $\mathcal{P}_R$, and vice versa.
To this end, we build the feature augmentation sub-network as a stack of alternating self- and cross-attention layers, interleaved with normalization layers.
The architecture of the attention layers is depicted in Figure~\ref{fig:attention}, implementing a residual block with message passing for feature update.
We set the feature-augmentation network up as a stack of fully-connected graph-attention layers, thereby letting the network learn which connections are relevant for the current point feature from all possible connections and to only attend to those via Multi-Head Softmax-Attention.
This allows to embed information of the relevant topology from both within and between point-sets in an iterative fashion into the feature descriptors, resulting in two sets $\hat{\mathcal{F}}_S: \{\mathit{f}_i \in \mathbb{R}^d, i = 1, ..., M\}$ and $\hat{\mathcal{F}}_R: \{\mathit{f}_j \in \mathbb{R}^d, j = 1, ..., N\}$ of point features for matching.

\subsection{Feature Matching}
After feature augmentation, matching is done by calculating the similarity score matrix $\mathbf{S} \in \mathbb{R}^{M,N}$ between the point feature descriptors $\mathcal{F}^m_S$ and $\mathcal{F}^m_R$ of all possible point pairs $\mathbf{p}_{i,j} = \{p_i \in \mathcal{P}_S, p_j \in \mathcal{P}_R\}$ using dot-product similarity:

\begin{equation}
 \mathbf{S}: s_{i,j} = \left< \mathit{f}_i, \mathit{f}_j \right>.
\end{equation}

Since we are interested in finding point-correspondences, we interpret the optimization problem of equation~(\ref{eq:perm}) in terms of the optimal transport problem~\cite{OT}, using the similarity score $\mathbf{S}$ as its cost.
We find an approximate solution $\mathbf{C}^*$ by adding a row and column of slack variables to $\mathbf{S}$ as detailed in equation~\ref{eq:perm_relaxed} and applying a few iterations of the Sinkhorn-Algorithm as a differentiable approximation to the Hungarian Algorithm for the solution of optimal transport~\cite{SinkhornOT, hungarian, sinkhorn}.
Finally, we threshold the resulting approximate permutation matrix $\mathbf{C}^*$ by threshold $\mathit{t}_m \in [0, 1]$ and take mutual row- and column-wise maxima as point correspondences for calculation of the rigid transformation $\{\mathbf{R}, \mathbf{t}\}$ aligning the point sets using SVD.

\subsection{Loss}

As loss for network training we employ the binary cross entropy loss between the predicted permutation matrix $\mathbf{C}^*$ and the ground truth correspondence matrix $\mathcal{G}_{gt}$:

\begin{equation}
 \mathcal{L}_{BCE} = - \sum_{i,j} \mathit{g}_{i,j} \log \hat{c}_{i,j} + (1 - \mathit{g}_{i,j}) \cdot \log (1 -\hat{c}_{i,j}).
\end{equation}

\section{Experiments}
In order to evaluate the performance of our proposed registration method, we perform two experiments.
The first experiment~\ref{sec:mnet40} tests the performance on synthetic data of ModelNet40~\cite{modelnet} for different settings of noise and overlap.
The second experiment described in section~\ref{sec:dmv} tests the generalization ability using LiDAR point clouds of the Kitti Odometry Benchmark~\cite{kitti} and custom high-quality real-world object scans, using only models trained on synthetic data in the experiment of section~\ref{sec:mnet40}.

Throughout the experiments, we have chosen the following parameters for our network:
The feature dimension is chosen as $d=128$, the number of layers and layer dimensions in the feature encoder of the feature head follows the parameterization of DGCNN~\cite{dgcnn} with a neighbourhood size of $\mathcal{N} = 20$.
The location encoder is chosen as a 4 layer MLP with layer dimensions $[16, 32, 64, 128]$.
Our feature-augmentation graph-attention network consists of 9 stacks of consecutive self- and cross-attention layers with 2 attention heads.
For normalization, batch-norm is chosen throughout the network.
The number of Sinkhorn-iterations is set to $10$ for both, training and inference.
We train the network on a single registration iteration per example, testing is done with a second iteration, feeding the source point cloud aligned by the result of the first iteration again through the network.

Model training usually converges after training for two days using AdamW optimizer with learning rate $1e^{-4}$ on a Nvidia GeForce RTX3090 (between 800 and 1000 epochs).

\subsection{\label{sec:mnet40}Experiments on ModelNet40}
ModelNet40 consists of $12,311$ meshed CAD models in $40$ object categories, spanning a vast array of scales from chairs to airplanes.
Consistent with previous work, we use the pre-sampled point clouds provided by Shapenet~\cite{modelnet2048}, consisting of 2048 points per model to conduct the experiments.
For easy comparison we follow the setup of~\cite{RGM} and perform the same experiments.
All experiments with exemption of subsection~\ref{sec:unseen_crop} follow the official training and testing split, with an additional 80:20 split of the official training set for training and validation.
The experiment described in subsection~\ref{sec:unseen_crop} uses the first 20 object classes of the training set for training, the first 20 object classes of the test set for validation and the remaining 20 classes for testing.
The point clouds already come scaled to fit within the unit circle, therefor all measures related to point distance are given in a normalized scale.
As in~\cite{RGM}, we sample 1024 points at random from the point clouds and apply random rotations within $[0^\circ, 80^\circ]$ around a random axis and random translations within $[-0.5, 0.5]$ in normalized units.

Registration performance in measured using the same metrics as~\cite{RGM}, that is residual transformation errors of $\{\mathbf{R}, \mathbf{t}\}$ as mean isotropic errors (MIE) as proposed by~\cite{RPMNet}, as well as clipped chamfer distance (CCD) between reference point cloud $\mathbf{Y}$ and transformed source point cloud $\hat{\mathbf{X}}$ after registration:

\begin{equation}
 \begin{split}
  CCD(\hat{\mathbf{X}}, \mathbf{Y}) & = \sum_{\hat{x}_i \in \mathbf{\hat{X}}} \min(\underset{y_j \in \mathbf{Y}}{\min}(||\hat{x}_i - y_j ||^2_2), r) \\
  & + \sum_{\hat{y}_j \in \mathbf{Y}} \min(\underset{\hat{x}_i \in \mathbf{\hat{X}}}{\min}(||\hat{x}_i - y_j ||^2_2), r),
 \end{split}
\end{equation}
with clip distance $r = 0.1$.
Furthermore, we report registration recall (RR), defined as percentage of registration results with residual errors MAE$(\mathbf{R}) < 1^\circ$ and MAE$(\mathbf{t}) < 0.1$.
To keep consistent with previous research, we also state the residual transformation errors in terms of mean absolute errors (MAE) as proposed by~\cite{DCP}, which is anisotropic.
Errors related to rotations are given in degrees, errors related to distance are normalized to object size (since the data in ModelNet40 does not have a common scale and is normalized to the unit circle).

The design of our registration method provides us directly with information on the reliability and success of a matching attempt.
Using the value of the matching score $s_{i,j}$ matching point $p_i$ to point $p_j$ as well as the number of found matches, we can reject invalid registrations.
To this end, we provide results for matching thresholds $t_m = 0.5$ and rejecting registrations with less than $3$ correspondences.
Evaluation of registration errors is done on successful registrations only, stating the percentage of successful registrations in braces after the method name.
Registration recall for our method is provided with respect to the full number of examples in the testing set, thereby making it directly comparable.
In practical applications, failed registrations can easily be rectified by either performing batched registrations with different samplings for a single registration task or repeating the registration with a different subset of points in case of failure.
Please note that the main competing methods do not allow any insight like this without knowledge of the underlying true registration, since RPMNet~\cite{RPMNet} works on soft-correspondences, RGM~\cite{RGM} only provides hard correspondences without associated score and returned in our experiments always more than $3$ matches.
Results of the comparing methods are reproduced from~\cite{RGM}.

\subsubsection{Full and clean data}
The first experiment can be considered a baseline in registration performance, since the transformation has to be recovered from a full set of 1024 exact and noise-free correspondences and is mainly reproduced for completeness.
From Table~\ref{tab:mnet40_clean} we can see that basically all methods are able to almost perfectly register the point clouds with MAE$(\mathbf{R})$ below or around $1^\circ$.
Only ICP struggles in comparison.

\begin{table}[ht]
\caption{Registration performance on clean point clouds.}
 \begin{adjustbox}{width=\columnwidth}
 \def\arraystretch{1.2}
  \begin{tabular}{l|r r r r r r}
   method                & MIE(R) & MIE(t)  & MAE(R)& MAE(t)  & CCD     & RR       \\ \hline
   ICP~\cite{icp-o3d}    & 6.4467 & 0.05446 & 3.079 & 0.02442 & 0.03009 &  74.19 \% \\
   FGR~\cite{FGR}        & 0.0099 & 0.00010 & 0.006 & 0.00005 & 0.00019 &  99.96 \% \\
   IDAM~\cite{idam}      & 1.3536 & 0.02605 & 0.731 & 0.01244 & 0.04470 &  75.81 \% \\
   DeepGMR~\cite{deepgmr}& \textit{0.0156} & \textit{0.00002} & \textit{0.001} & \textit{0.00001} & \textit{0.00003} & \textbf{100.00} \% \\
   RPMNet~\cite{RPMNet}  & 0.2464 & 0.00112 & 0.109 & 0.00050 & 0.00089 &  98.14 \% \\
   RGM~\cite{RGM}        & \textbf{0.0103} & \textless\textbf{0.00001} & \textless\textbf{0.001} & \textless\textbf{0.00001}& \textless\textbf{0.00001}& \textbf{100.00} \% \\
   Ours ($100.00\%$)& 0.0150 & 0.00009 & 0.007 & 0.00004 & 0.00014 & 99.92 \% \\
  \end{tabular}
 \end{adjustbox}
 \label{tab:mnet40_clean}
\end{table}

\subsubsection{\label{sec:noise}Additive gaussian noise}
In this experiment, source and reference point sets are sampled independently, so only a few perfect correspondences may exist.
Additionally, we add gaussian noise sampled from $\mathcal{N}(0, 0.01)$ and clipped to the range $[-0.05, 0.05]$ to the point locations independently, thereby eliminating all perfect correspondences.
Point correspondences and point normals are then re-established, following the procedure of~\cite{RGM}, first finding mutual nearest neighbours and then adding remaining nearest neighbours, all within a maximum distance of 0.05 between corresponding points.

As can be expected, the performance degrades to a certain degree.
The results listed in Table~\ref{tab:mnet40_noise} show that the learning based methods still hold up rather well with MAE$(\mathbf{R})$ around or below $3^\circ$.
RPMNet, RGM as well as our method still achieve a RR of more than $90\%$.
Interestingly, the performance of ICP does not degrade, showing its robustness to outliers.

\begin{table}[h]
\caption{Registration performance with additive gaussian noise.}
 \begin{adjustbox}{width=\columnwidth}
 \def\arraystretch{1.2}
  \begin{tabular}{l|r r r r r r}
   method                &  MIE(R) & MIE(t)  & MAE(R)& MAE(t)  & CCD     & RR      \\ \hline
   ICP~\cite{icp-o3d}    &  6.5030 & 0.04944 & 3.127 & 0.02256 & 0.05387 & 77.59 \% \\
   FGR~\cite{FGR}        & 10.0079 & 0.07080 & 5.405 & 0.03386 & 0.06918 & 30.75 \% \\
   IDAM~\cite{idam}      &  3.4916 & 0.02915 & 1.818 & 0.01516 & 0.05436 & 49.59 \% \\
   DeepGMR~\cite{deepgmr}&  2.2736 & 0.01498 & 1.178 & 0.00716 & 0.05029 & 56.32 \% \\
   RPMNet~\cite{RPMNet}  &  \textit{0.5773} & \textit{0.00532} & \textit{0.305} & \textit{0.00253} & \textit{0.04257} & \textit{96.68} \% \\
   RGM~\cite{RGM}        &  \textbf{0.1496} & \textbf{0.00141} & \textbf{0.080} & \textbf{0.00069} & \textbf{0.04185} & \textbf{99.51} \% \\
   Ours ($98.82\%$)      &  0.8560 & 0.00635 & 0.518 & 0.00296 & 0.04297 & 93.64 \% \\
  \end{tabular}
 \end{adjustbox}
 \label{tab:mnet40_noise}
\end{table}

\subsubsection{\label{sec:crop}Registration of noisy, partially overlapping sets}
In this experiment, in addition to additive gaussian noise, both source and reference point clouds are independently cropped along a random plane to $70\%$ of their original size, resulting in variable overlap of at least $40\%$. This experimental setup corresponds closest to general real-world applications.
From Table~\ref{tab:mnet40_crop} we see that, with exception of RPMNet~\cite{RPMNet}, RGM~\cite{RGM} and ours, the registration performance degrades beyond anything what can be deemed usable in any applications.
Notably, the registration performance on recovered registrations of our method is the same as in the previous experiment with full overlap, albeit losing in successful registrations and in registration recall.
Comparing the registration recall of $77.2\%$ to the percentage of recovered registrations of $84.3\%$, we see the merit of our architecture and the ability to predict whether a registration attempt was successful.

\begin{table}[h]
\caption{Registration performance with only partial overlap and additive gaussian noise.}
 \begin{adjustbox}{width=\columnwidth}
 \def\arraystretch{1.2}
  \begin{tabular}{l|r r r r r r}
   method                & MIE(R)  & MIE(t)  & MAE(R) & MAE(t)  & CCD     & RR [\%] \\ \hline
   ICP~\cite{icp-o3d}    & 24.8777 & 0.26685 & 12.456 & 0.12465 & 0.11511 &  6.56 \% \\
   FGR~\cite{FGR}        & 42.4292 & 0.30214 & 23.185 & 0.14560 & 0.12118 &  5.23 \% \\
   IDAM~\cite{idam}      & 16.9724 & 0.19209 &  8.905 & 0.09192 & 0.12393 &  0.81 \% \\
   DeepGMR~\cite{deepgmr}& 70.9143 & 0.45705 & 43.683 & 0.22479 & 0.14401 &  0.08 \% \\
   RPMNet~\cite{RPMNet}  &  1.6985 & 0.01763 &  0.864 & 0.00834 & 0.08457 & \textit{80.59} \% \\
   RGM~\cite{RGM}        &  \textit{0.9298} & \textit{0.00874} &  \textit{0.492} & \textit{0.00414} & \textit{0.08238} & \textbf{93.31} \% \\ 
   Ours ($84.32\%$)      &  \textbf{0.8854} & \textbf{0.00721} &  \textbf{0.484} & \textbf{0.00347} & \textbf{0.08119} & 77.19 \% \\
  \end{tabular}
 \end{adjustbox}
 \label{tab:mnet40_crop}
\end{table}

\subsubsection{\label{sec:unseen_crop}Partial overlap of unseen object categories}
The difference to the experiment outlined in section~\ref{sec:crop} is that now we only train on the first 20 object categories of ModelNet40, but evaluate on the remaining 20 categories.
Thereby we can explore to what extent the learned registration networks are able to generalize to geometries not present in training.
An interesting fact evident in the results listed in Table~\ref{tab:mnet40_crop_unseen} is that the performance of all methods except RPM~\cite{RGM} does not decline much relative to the experiment done on known categories, whereas RPM almost doubles its residual errors.
Although very powerful in establishing good correspondences, the neural network architecture in RPM seems to learn geometries by heart, hampering its generalization ability, whereas our method performs as strong as it did before, outperforming RGM in all measures.
This again exemplifies the merit of feature augmentation at test time for optimal matching success.
Furthermore we would like to point out that although our method is not able to successfully register all examples in the first attempt, using the match threshold $t_m$ and the number of found matches, we can precisely predict unsuccessful attempts.
In all experiments, RR is close to the number of valid examples within a margin of about $5\%$.

\begin{table}[h]
\caption{Registration performance on unseen categories, partial overlap and gaussian noise.}
 \begin{adjustbox}{width=\columnwidth}
 \def\arraystretch{1.2}
  \begin{tabular}{l|r r r r r r}
   method                & MIE(R)  & MIE(t)  & MAE(R) & MAE(t)  & CCD     & RR      \\ \hline
   ICP~\cite{icp-o3d}    & 26.6447 & 0.27774 & 13.326 & 0.13033 & 0.11879 &  6.71 \% \\
   FGR~\cite{FGR}        & 41.9631 & 0.29106 & 23.950 & 0.14067 & 0.12370 &  5.13 \% \\
   IDAM~\cite{idam}      & 19.3249 & 0.20729 & 10.158 & 0.10063 & 0.12921 &  0.95 \% \\
   DeepGMR~\cite{deepgmr}& 71.0677 & 0.44632 & 44.363 & 0.22039 & 0.14728 &  0.24 \% \\
   RPMNet~\cite{RPMNet}  &  1.9826 & 0.02276 &  1.041 & 0.01067 & 0.08704 & 75.59 \% \\
   RGM~\cite{RGM}        &  \textit{1.5457} & \textit{0.01418} &  \textit{0.837} & \textit{0.00674} & \textit{0.08469} & \textit{84.28} \% \\
   Ours (89.10\%)        &  \textbf{0.8695} & \textbf{0.00871} &  \textbf{0.434} & \textbf{0.00432} & \textbf{0.08299} & \textbf{85.78} \% \\
  \end{tabular}
 \end{adjustbox}
 \label{tab:mnet40_crop_unseen}
\end{table}

\subsection{\label{sec:dmv}Generalization to real-world 3D scans}
For real-world application, the ability of 3D registration methods to generalize to new and different geometries as well as capturing modalities is crucial.
To this end, we compare the registration performance of the best performing methods trained on ModelNet40 as detailed in section~\ref{sec:mnet40} on two datasets, a custom dataset (publication is planned) as well as the the well known Kitti Dataset~\cite{kitti}.
The custom dataset consist of objects scans of 10 objects taken with an Artec Leo~\cite{ArtecLeo} handheld 3D scanner, for each object up to 10 overlapping partial scans exist, with between 10.000 and 50.000 points each.
Figure~\ref{fig:dmv} shows a registration example of this dataset.
Transformations are generated within the same constraints as in the experiments on ModelNet40.
We report registration accuracy in terms of MIE$(\mathbf{R})$, MIE$(\mathbf{t})$ and registration recall.
Since the objects in this dataset have a common scale, MIE$(\mathbf{t})$ is reported in millimeters, registration recall is defined as percentage of registration results with residual errors MIE$(\mathbf{R}) < 1^\circ$ and MIE$(\mathbf{t}) < 5 mm$.
Again, the number of in brackets behind versions of our method states the respective percentage of \emph{valid} registrations.
For the experiments on Kitti, we follow the established praxis~\cite{geotrans,FCGF,predator} of testing on sequences 8-10, testing registration performance of point cloud pairs at least 10m apart.
As in~\cite{geotrans,FCGF,predator}, we use ground truth poses refined by ICP, MIE$(\mathbf{t})$ is reported in meters, and registration recall is defined as percentage of registration results with residual errors MIE$(\mathbf{R}) < 5^\circ$ and MIE$(\mathbf{t}) < 2 m$.
Note that for fairness we applied an additional data normalization step for RPM-Net and RGM, scaling the data to fit into the unit circle for registration, thus making the input points span the same range as the training data of ModelNet40.
From Table~\ref{tab:dmv} we can see that our algorithm generalizes well to high quality 3D scans, the models trained on partial overlapping data outperform both RGM~\cite{RGM} and RPMNet~\cite{RPMNet} by a large margin in all metrics.
For registration of large-scale outdoor scenes of Kitti, a domain-gap for all methods is noticeable.
Nonetheless, our method still performs reasonably well given the circumstances, with registration recall of around $50\%$ and mean errors of $3.1^\circ$ and $3.5m$ for the best generalizing models trained with only partial overlap, again showing its robustness to different data modalities.
Furthermore, the strong ability to predict which registrations were successful is visible from comparing the number of $51.1\%$ valid registrations to the RR of $49.7\%$ for the model trained on unseen categories.
Again, we can observe that while a powerful registration method, RGM seems to overfit on the training modalities, being beaten even by RPM-Net trained for the experiment on noisy data and unseen categories, whereas our method is rather robust to changes in sampling, overlap and geometry.
Interestingly, for both, RGM and RPM-Net, models trained on the harder cases of only partial overlap often lead to a decrease in generalization performance, whereas our methods ability to generalize to different data improves with the difficulty of the training task.

\begin{table}[h]
\caption{Generalization to real world objects scanned using a handheld 3D scanner, using the models trained on ModelNet40 from the experiments in section~\ref{sec:mnet40}.
Here, the name in the column \emph{experiment} refers to the experiment in which the method was trained, no further data augmentation has been done besides random sub-sampling.
For RR, thresholds are set as MIE$(\mathbf{R}) < 1^\circ$ and MIE$(\mathbf{t}) < 5 mm$.}
 \begin{adjustbox}{width=\columnwidth}
 \def\arraystretch{1.2}
  \begin{tabular}{l| l |r r r}
   method        & experiment & MIE(R) [$^\circ$]  & MIE(t) [mm] & RR [\%] \\ \hline
   \multirow{3}{*}{RPMNet~\cite{RPMNet}}
    & clean  & 23.9 & 88.2 & 0.8 \% \\
    & noise  &  1.8 &  6.2 & 62.8 \% \\
    & unseen &  4.4 & 14.9 & 65.0 \% \\
   \hline
   \multirow{4}{*}{RGM~\cite{RGM}}
    & clean  & 6.1 & 22.8 & 10.5 \% \\
    & noise  & 3.3 & 12.3 & 32.5 \% \\
    & crop   & 5.8 & 24.8 & 25.4 \% \\
    & unseen & 7.0 & 26.6 & 26.0 \% \\
   \hline
   \multirow{4}{*}{Ours}
    & clean ($100.0\%$) &  6.0 & 22.4 & 5.0 \% \\
    & noise  ($97.6\%$) &  1.2 & 5.0 & 58.6 \% \\
    & crop  ($99.1\%$)  & \textit{0.6} & \textit{1.9} & \textit{74.3} \% \\
    & unseen ($98.7\%$) & \textbf{0.6} & \textbf{1.7} & \textbf{76.7} \% \\
  \end{tabular}
 \end{adjustbox}
 \label{tab:dmv}
\end{table}

\begin{table}[ht]
\caption{Generalization to data from Kitti Odometry Benchmark, again using the models trained on ModelNet40 from the experiments in section~\ref{sec:mnet40}.
For RR, thresholds are set as MIE$(\mathbf{R}) < 5^\circ$ and MIE$(\mathbf{t}) < 2 m$.
}
 \begin{adjustbox}{width=\columnwidth}
 \def\arraystretch{1.2}
  \begin{tabular}{l| l |r r r}
   method        & experiment & MIE(R) [$^\circ$]  & MIE(t) [m] & RR [\%] \\ \hline
   \multirow{3}{*}{RPMNet~\cite{RPMNet}}
    & clean  & 100.1 & 9.8 & 0.0 \% \\
    & noise  & 5.6 &  8.3 & 0.5 \% \\
    & unseen & 4.7 & 7.5 & 0.7 \% \\
   \hline
   \multirow{4}{*}{RGM~\cite{RGM}}
    & clean  & 6.5 & 9.4 & 0.0 \% \\
    & noise  & 6.0 & 8.1 & 0.2 \% \\
    & crop   & 6.7 & 8.4 & 0.7 \% \\
    & unseen & 9.2 & 8.7 & 0.0 \% \\
   \hline
   \multirow{4}{*}{Ours}
    & clean ($100.0\%$) & 7.2 & 9.6 & 0.0 \% \\
    & noise ($88.8\%$)& 14.4 & 10.3 & 1.6 \% \\
    & crop ($59.1\%$) & \textit{3.4} & \textit{3.5} & \textit{47.0} \% \\
    & unseen ($51.1\%$) & \textbf{3.1} & \textbf{3.5} & \textbf{49.7} \% \\
  \end{tabular}
 \end{adjustbox}
 \label{tab:kitti}
 \vspace{-0.25cm}
\end{table}

\subsection{Resource Consumption and performance}
Registration performance is not the only relevant criterion for the usability of an algorithm.
Execution time as well as compute resource needs are limited especially in mobile applications and are therefor a further relevant measure in algorithm selection.
To this end, we compare our algorithm in terms of complexity and resource needs to the two best competing methods.
Model complexity is measured in the number of trainable parameters.
Compute resource needs are given in GB of GPU memory use for batch sizes of 20, 5, and 1, as well as registration speed measured in registrations per second.
We can see from Table~\ref{tab:resource}, that our method is both more light-weight and faster while still providing competitive results.

\begin{table}[h]
\caption{Resource consumption of the best performing methods on a Nvidia GeForce RTX3090. Memory use is provided for batch sizes 20, 5, and 1.}
 \begin{adjustbox}{width=\columnwidth}
 \centering
 \def\arraystretch{1.2}
  \begin{tabular}{l|r r r r r}
   method              & param [\#]  & mem@20 & mem@5 & mem@1 & rate [\#/s] \\ \hline
   RPMNet~\cite{RPMNet}& $0.91e^6$ & $6.50$ GB & $3.2$ GB & $2.2$ GB & $45.9$ \\
   RGM~\cite{RGM}      & $25.0e^6$ & $7.20$ GB & $3.4$ GB & $2.4$ GB & $ 6.6$ \\
   Ours                & $ 4.4e^6$ & $4.47$ GB & $2.6$ GB & $2.2$ GB & $62.0$ \\
  \end{tabular}
 \end{adjustbox}
 \label{tab:resource}
 \vspace{-0.25cm}
\end{table}

\subsection{Ablation Study}
In order to evaluate the benefit of different parts in our feature head, we test the following configurations.
Networks are trained on the task of partial overlap, as in section~\ref{sec:crop} using the same random seed, with the following architectural differences:
\begin{itemize}
 \item Location encoder: the feature head only uses the location encoder.
 \item Feature only: the feature head only uses the local point feature network.
 \item additive fusion: the MLP fusing position encoding and local point feature is replaced by a simple addition of feature vectors.
 \item MLP fusion: this is the full network architecture, consisting of the feature head with location encoder, point feature network and MLP for feature fusion.
\end{itemize}

Please note that the networks have not been trained to full convergence, since only a qualitative difference is required.
For testing, the same modalities as for the experiments in section~\ref{sec:mnet40} have been employed.
From the results in Table~\ref{tab:ablation} we can see, that each additional structure improves the overall performance, the method works best if we let the network learn how to combine both feature vectors.

\begin{table}[h]
\caption{
Ablation study testing the influence of the different parts of our feature head. The full head with both, local feature encoder and position encoder fused by an small MLP, performs best.
Note that the neural networks were not trained to full convergence in this study.}
 \centering
 \def\arraystretch{1.2}
  \begin{tabular}{l|r r r}
   variant          & MIE(R) [$^\circ$] & MIE(t) & RR [\%] \\ \hline
   location encoder & 2.90 & 0.021 & 72.5 \% \\
   feature encoder  & 1.99 & 0.016 & \textit{78.9} \% \\
   additive fusion  & \textit{1.76} & \textit{0.014} & 77.0 \% \\
   MLP fusion       & \textbf{1.29} & \textbf{0.012} & \textbf{79.0} \% \\
  \end{tabular}
 \label{tab:ablation}
\end{table}

\section{Conclusion}
In this paper, we presented GAFAR, a novel, light-weight algorithm for point set registration using an end-to-end learnable deep neural network for feature encoding and correspondence prediction.
Its performance is competitive while being faster and less demanding on resources compared to other state-of-the-art methods, which makes it well suited for applications with constraints on compute resources, power consumption and runtime.
Our method shows very high generalization capability to different data modalities and exhibits little overfit to geometry details of the training set.
The strong performance for partial overlap, even for object classes not present in training, shows the merits of the cross-attention mechanism for feature augmentation.
A further benefit of our method is its ability to provide an indication on the quality of predicted correspondences, thereby giving opportunity to tune between high registration accuracy and high recall as well as to reject failed or bad registrations without additional knowledge.
In practice, failure cases can be remedied by either performing multiple registrations with different sub-sampling in parallel in a batched fashion, or by repeating the registration with a different sample in case of failure.

In the future, we plan to tackle the limitation to only small subsets of point clouds by applying the underlying architectural principles to the registration of large point sets directly, while still keeping with the paradigm of light-weight architecture and fast execution.

\addtolength{\textheight}{-0.5cm}   



%

\section*{Acknowledgment}
This work was supported by Land Steiermark within the research initiative ``Digital Material Valley Styria''.


\bibliographystyle{bib/IEEEtran} 
\bibliography{bib/IEEEfull,bib/references}

\end{document}